%% file: final.tex
\documentclass{article}

\usepackage{microtype}
\usepackage{graphicx}
\usepackage{subfigure}
\usepackage{booktabs} 
\usepackage{stfloats}
\usepackage[justification=centering]{caption}
\usepackage{enumitem}

\usepackage{hyperref}

\usepackage[accepted]{icml2023}

\usepackage{amsmath}
\usepackage{amssymb}
\usepackage{mathtools}
\usepackage{amsthm}
\makeatletter
\def\th@plain{%
  \thm@notefont{}
  \itshape 
}
\def\th@definition{%
  \thm@notefont{}
  \normalfont 
}
\makeatother

\usepackage[capitalize,noabbrev]{cleveref}

\theoremstyle{plain}
\newtheorem{theorem}{Theorem}[section]

\newtheorem{lemma}[theorem]{Lemma}

\theoremstyle{definition}

\theoremstyle{remark}

\usepackage[textsize=tiny]{todonotes}

\usepackage{nicefrac}
\usepackage{bm}
\usepackage{xspace}
\usepackage{xcolor}
\usepackage{amsmath,amssymb,amsthm,amsfonts}

\usepackage{babel,blindtext}
\usepackage{epstopdf}
\usepackage{booktabs}
\usepackage{multirow}
\usepackage{diagbox}       
\usepackage{microtype}

\usepackage{courier}
\usepackage{newfloat}
\usepackage{listings}
\newcommand{\vct}[1]{\bm{#1}}
\newcommand{\scheme}{FedVS\xspace}

\let\emptyset\varnothing

\icmltitlerunning{FedVS: Straggler-Resilient and Privacy-Preserving Vertical Federated Learning for Split Models}

\begin{document}

\twocolumn[
\icmltitle{FedVS: Straggler-Resilient and Privacy-Preserving Vertical Federated Learning for Split Models}



\icmlsetsymbol{equal}{*}

\begin{icmlauthorlist}
\icmlauthor{Songze Li}{1,2}
\icmlauthor{Duanyi Yao}{2}
\icmlauthor{Jin Liu}{1}
\end{icmlauthorlist}

\icmlaffiliation{1}{The Hong Kong University of Science and Technology (Guangzhou)}
\icmlaffiliation{2}{The Hong Kong University of Science and Technology}

\icmlcorrespondingauthor{Songze Li}{songzeli@ust.hk}

\icmlkeywords{Machine Learning, ICML}

\vskip 0.3in
]



\printAffiliationsAndNotice{}  

\begin{abstract}
  In a vertical federated learning (VFL) system consisting of a central server and many distributed clients, the training data are vertically partitioned such that different features are privately stored on different clients. The problem of split VFL is to train a model split between the server and the clients. This paper aims to address two major challenges in split VFL: 1) performance degradation due to straggling clients during training; and 2) data and model privacy leakage from clients' uploaded data embeddings. We propose FedVS to simultaneously address these two challenges. The key idea of FedVS is to design secret sharing schemes for the local data and models, such that information-theoretical privacy against colluding clients and curious server is guaranteed, and the aggregation of \emph{all} clients' embeddings is reconstructed losslessly, via decrypting computation shares from the non-straggling clients. Extensive experiments on various types of VFL datasets (including tabular, CV, and multi-view) demonstrate the universal advantages of FedVS in straggler mitigation and privacy protection over baseline protocols. 
\end{abstract}

\input{intro_modi}

\input{setting}

\input{preliminary}
\input{protocol}
\input{analysis}
\input{experiment}
\input{conclusion}


\bibliography{reference}
\bibliographystyle{icml2023}

\newpage
\appendix
\onecolumn
\input{appendix.tex}

\end{document}

%% file: intro_modi.tex
\section{Introduction}

Federated learning (FL)~\cite{mcmahan2017communication,ZHANG2021106775} is an emerging machine learning paradigm where multiple clients (e.g., companies) collaborate to train a machine learning model while keeping the raw data decentralized. 
Based on how data is partitioned across clients, FL can be categorized into horizontal FL and vertical FL. In horizontal FL (HFL), each client possesses a distinct set of data samples who share the same set of features; in vertical FL (VFL), each client has a distinct subset of features for a collection of shared samples. While current FL research largely focused on HFL, VFL is attracting more attention due to its suitability for enabling data augmentation for a wide range of applications in decision making~\cite{cheng2021secureboost}, risk control~\cite{cheng2020federated}, and health care~\cite{lee2018privacy}. In a basic VFL setting (see, e.g.,~\cite{DBLP:journals/corr/abs-1911-09824,DBLP:journals/corr/abs-2001-11154}), the FL system trains a local model for each client, which are jointly utilized to perform inferences. A more general VFL setting, named split VFL~\cite{DBLP:journals/corr/abs-2008-04137}, incorporates the idea of split learning~\cite{DBLP:journals/corr/abs-1812-00564}, and jointly trains a central model at the server and local models at the clients. 


In a training round of split VFL, all clients forward propagate their local data using local models, and send the output embeddings to the server; the server then aggregates these embeddings and continues forward prorogation through its central model. Having computed the loss, the server back propagates to update the central model, and sends the gradients of the embeddings to the clients to update the local models. An ideal round requires synchronous aggregation of clients' embeddings. 
However, this is severely challenged by the system and task heterogeneity commonly observed in VFL, which is caused by variability of clients' storage, computation and communication resources, and local data and model complexities~\cite{reisizadeh2022straggler,wei2022vertical}.
Clients with slowest speeds of forward propagation, which we call \emph{stragglers}, become the bottleneck in training process, and cause detrimental effects on model convergence. 

One way to deal with stragglers is simply ignoring them, which however leads to slow convergence and model bias. Asynchronous VFL protocols have been proposed to enable asynchronous submissions of embeddings and model updates without client coordination~\cite{DBLP:journals/corr/abs-2007-06081,hu2019fdml}. However, this causes staleness of model updates that can degrade model performance. 
Under the synchronous framework, Flex-VFL~\cite{castiglia2022flexible} was proposed to enable flexible numbers of local model updates across clients, mitigating the slowdown of convergence caused by stragglers.
 
Other than stragglers, another key challenge for split VFL is privacy leakage through clients' embeddings. Various inference attacks have been developed 
to recover clients' private data and model parameters, from the uploaded raw embeddings (\cite{erdogan2021unsplit,jin2021cafe,li2021survey,luo2021feature,fu2022label}). 
Differential privacy (DP) has been adopted to defend inference attacks, which adds a DP noise layer on raw embeddings to protect data privacy (see, e.g.,~\cite{thapa2022splitfed,DBLP:journals/corr/abs-2007-06081,xu2021achieving}).
However, the added noises cause inaccurate computations of gradients, which subsequently leads to performance loss. Homomorphic encryption (HE) has also been utilized in VFL to protect embedding privacy, such that ciphertexts of embeddings are aggregated and only the summation of all embeddings is revealed~\cite{hardy2017private,yang2019parallel,cai2022secure}. 
These methods provide privacy for clients' data but cannot mitigate stragglers effectively. Recently in~\cite{shi2022efficient}, it is proposed to use secure aggregation~\cite{bonawitz2017practical} for privacy protection in asynchronous training of linear and logistic regression models over vertically partitioned data, which is nevertheless faced with slow convergence from asynchronous model updates.
Given the above challenges and the prior works, we ask the following question:

\textit{Can one design a synchronous split VFL protocol that is simultaneously lossless against unknown stragglers and provably private against curious server and clients?}
 
We answer this question in affirmative, via proposing a straggler-resilient and privacy-preserving split VFL protocol named \scheme. The key idea 
is to secret share local data and model of each client with peer clients, creating data redundancy across the network without any privacy leakage. Specifically, Lagrange Coded Computing (LCC)~\cite{yu2019lagrange} is adopted 
to improve computation and communication efficiencies. Averaging is chosen as the embedding aggregation method, such that the server only recovers the summation of the embeddings without knowing individual values. Clients utilize polynomial networks~\cite{livni2014computational} as local models, 
such that embedding summation can be \emph{losslessly} reconstructed at the server using polynomial interpolation. Leveraging the threshold property of polynomial interpolation, computation results from only a subset of clients are needed, effectively mitigating the stragglers. We theoretically analyze the straggler resilience and privacy guarantees of \scheme, its convergence performance, and operational complexities.

We experimentally demonstrate the advantages of \scheme in straggler mitigation and privacy protection for split VFL systems. Over a wide range of tabular, computer vision, and multi-view datasets, \scheme uniformly achieves the fastest convergence and highest accuracy, over baselines with or without privacy protection. The impacts of design parameters of \scheme on its performance and privacy are also empirically studied.


\subsection*{Related works}

\textbf{Straggler-resilient FL:}

\textit{Horizontal FL}: 
Proposed in~\cite{reisizadeh2022straggler}, FLANP starts the training with server exchanging models with a group of fast-responding clients, 
and gradually involves the slower clients. Sageflow~\cite{park2021sageflow} proposes to group the local models from stragglers according to their staleness, and aggregate the models from different groups with appropriate weights. 
In \cite{dhakal2019coded,prakash2020coded,9834445,sun2022stochastic}, clients share a part of their local data with the server, who computes the missing results from stragglers;
while in~\cite{schlegel2021codedpaddedfl,shao2022dres}, clients secret share their data with each other and perform local training on shares of all clients, such that the server losslessly decodes the gradient over all clients' data from only a subset of non-straggling clients.  
On the other hand, many asynchronous HFL protocols \cite{xie2019asynchronous,van2020asynchronous,li2021stragglers,huba2022papaya,chai2021fedat,nguyen2022federated} have been proposed to handle the straggler problem. 


\textit{Vertical FL}: For mitigating stragglers in VFL, 
Multiple asynchronous VFL protocols (see, e.g.,~\cite{DBLP:journals/corr/abs-2007-06081,9463409,zhang2021secure,li2020efficient,shi2022efficient,hu2019fdml}) have been proposed to reduce the waiting time for stragglers. 
VAFL~\cite{DBLP:journals/corr/abs-2007-06081} is designed for clients with intermittent connectivities, where each client individually updates its local model once connected with the server. DP is introduced to protect the privacy of local embeddings in VAFL, which nevertheless incurs performance loss. 
AFSGD-VP~\cite{9463409} is designed for the scenario where there is no central server and labels are held by multiple clients. It allows asynchronous data collection and model updating for label holders, and at the same time protects embedding privacy via a tree-structured aggregation scheme. 
AMVFL~\cite{shi2022efficient} proposes asynchronous aggregation to compute gradients, for linear and logistic regression problems, where local embeddings are protected by secret shared masks. 

\textbf{Privacy-preserving FL:} Current approaches to provide privacy protection for FL can be categorized into three types, which are homomorphic encryption (HE), DP, and secure multi-party computation (MPC)~\cite{liu2022privacy}. HE methods are applied to encrypt the local updates sent to the server (see, e.g.,\cite{chai2020secure,254465,cai2022secure}). It allows certain computations (e.g., addition) directly on the ciphertexts and noise-free recovery of computation results. However, the encryption and decryption introduce significant overheads. 
Compared with HE, DP is more efficient to provide privacy by injecting noises to the private data~\cite{wei2020federated,truex2020ldp,thapa2022splitfed,wang2020hybrid}. Nevertheless, the performance and convergence rate of the model suffer from the inaccurate computation results~\cite{dp:2,ddgmfl}. MPC protocols based on Shamir secret sharing have been proposed to securely aggregate clients' local models in HFL, such that the server learns nothing beyond the aggregated model~\cite{bonawitz2017practical,so2021turbo,bell2020secure,choi2020communication,so2022lightsecagg,liu2022efficient,9834750,jahani2022swiftagg+}. These protocols guarantee information-theoretic privacy for clients' local data, in the presence of client dropouts. 
Compared with these works, the proposed \scheme is the first MPC-based synchronous VFL protocol that simultaneously achieves information-theoretic privacy for each client's local data and model. Furthermore, in contrast to recovering model aggregation of non-straggling clients, \scheme achieves straggler resilience with no performance loss, i.e., the recovered embedding aggregation contains the local embeddings of all stragglers.

%% file: setting.tex
\begin{figure*}[t]
\minipage{0.33\textwidth}
  \centering
\includegraphics[width=\linewidth]{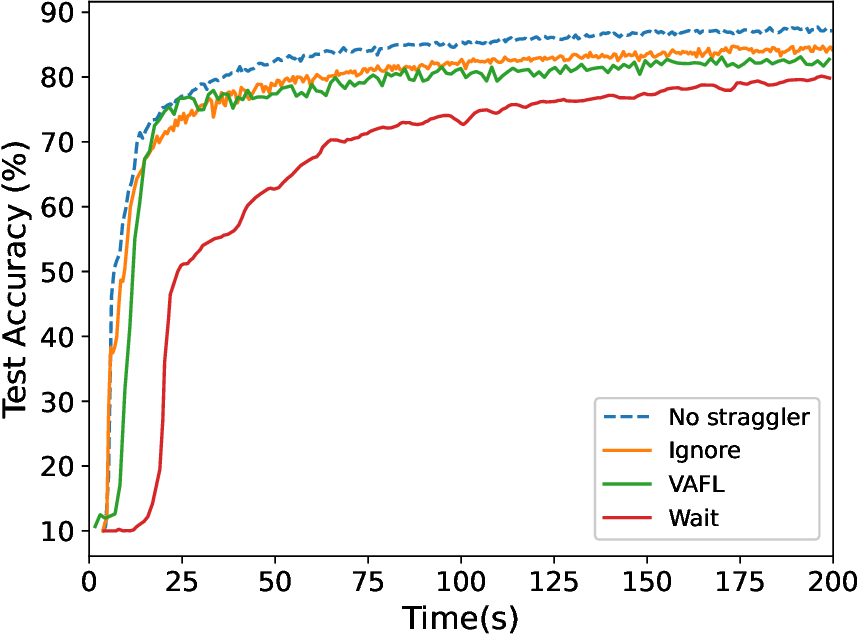}
  \subfigure{\small{(a) Concatenation}}
\endminipage\hfill
\minipage{0.33\textwidth}
  \centering
  \includegraphics[width=\linewidth]{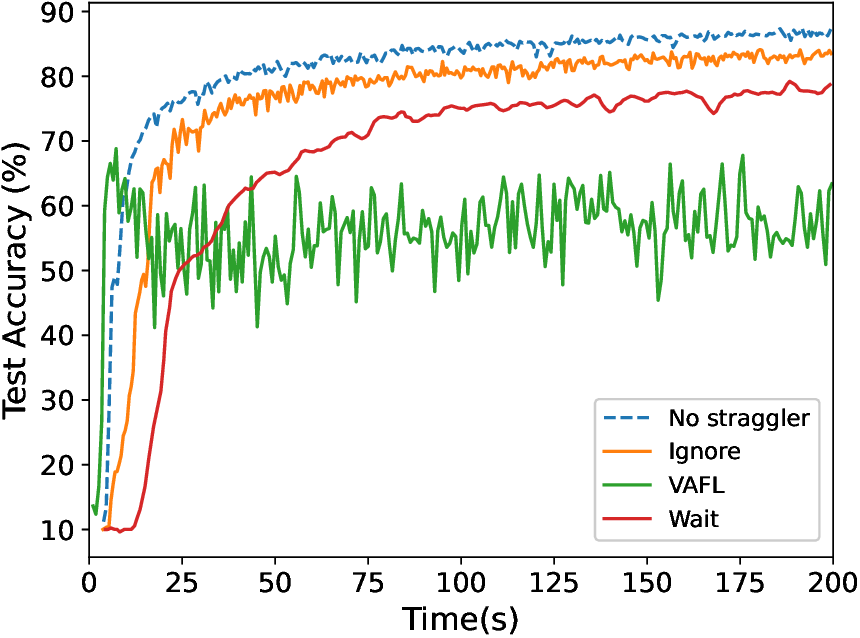}
  \subfigure{\small{(b) Element-wise average}}
\endminipage\hfill
\minipage{0.33\textwidth}%
  \centering
  \includegraphics[width=\linewidth]{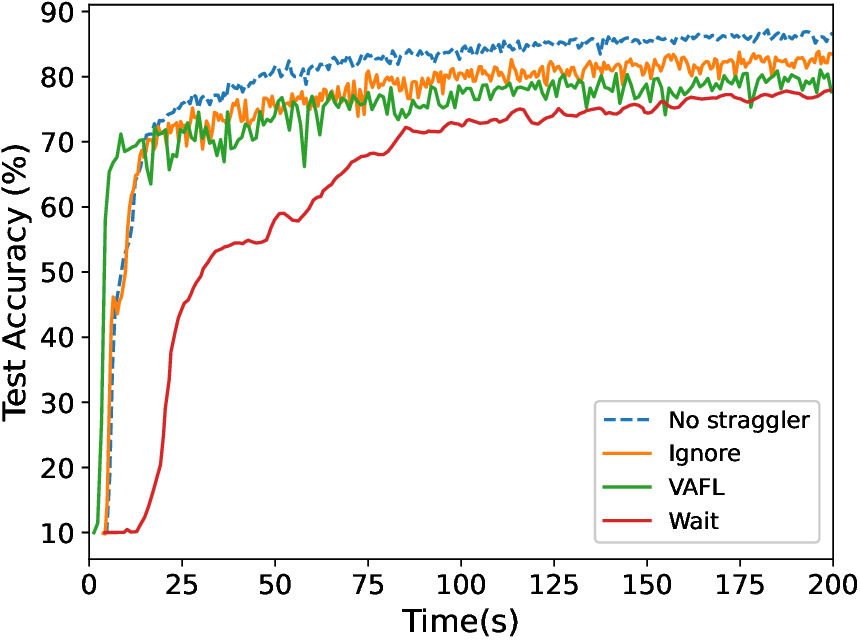}
  \subfigure{\small{(c) Element-wise maximum}}
\endminipage
\vspace{-3mm}
\caption{Test accuracies using different embedding aggregation methods and straggler handling strategies. 
}
\label{fig:pre}
\vspace{-5mm}
\end{figure*}

\section{Background and Motivations}\label{sec:setting}

\subsection{Split vertical federated learning} 
We consider a vertical federated learning (VFL) system that consists of a central server and $N$ clients. The training dataset  ${\cal S} = \{(\vct{x}^{(m)},\vct{y}^{(m)})\}_{m=1}^M$ contains $M$ input-label pairs, where each input $\vct{x}^{(m)} \in \mathbb{R}^d$ has $d$ features. The training set is vertically partitioned such that each client $n$ locally has a disjoint subset of $d_n$ features of each input. 
All labels 
are stored at the server. The VFL system aims to train a neural network that is split among server and clients. The server has a central model with parameters $\vct{W}_0$, and each client $n$ has a local model with parameters $\vct{W}_n$. Models on different clients may have different architectures, and hence the model parameters may have different dimensions. 

The server and clients collaboratively train their models to minimize the empirical loss $L((\vct{W}_n)_{n=0}^N; {\cal S}) \!=\!  \frac{1}{M} \sum_{m=1}^M \ell \left((\vct{W}_n)_{n=0}^N; (\vct{x}^{(m)},\vct{y}^{(m)})\right)$, for some loss function $\ell$.
The training is carried out via forward-backward prorogation over split models. In each round, for a batch ${\cal B}$ of $b$ inputs $\vct{X}^{({\cal B})} \in \mathbb{R}^{b \times d}$, we denote the partition at client $n$ as $\vct{X}_n^{({\cal B})} \in \mathbb{R}^{b \times d_n}$, for all $n \in [N] \triangleq \{1,\ldots,N\}$. To start, 
each client $n$ computes an embedding matrix $\vct{H}_n^{(\cal B)} \in \mathbb{R}^{b \times h_n}$, for some embedding dimension $h_n$, using its local network as $\vct{H}_n^{(\cal B)} = g_n(\vct{X}_n^{({\cal B})},\vct{W}_n)$, and sends it to the server. 
The server aggregates embeddings from all clients into a global embedding $\vct{H}^{({\cal B})}$. As discussed in~\cite{DBLP:journals/corr/abs-2008-04137}, the aggregation can be done in multiple ways, including concatenation, element-wise
average, 
and element-wise maximum. 
Next, the server feeds $\vct{H}^{({\cal B})}$ into the central network until the loss function $L$ is computed with the corresponding labels $\vct{Y}^{({\cal B})}$. 
In the backward propagation, the server computes the gradient $\nabla_{\boldsymbol{W}_0} L $ to update the central model with learning rate $\eta_0$, i.e., $\boldsymbol{W}_0 = \boldsymbol{W}_0 - \eta_0\nabla_{\boldsymbol{W}_0} L$. Then, for each $n \in [N]$,  the server computes the gradient $\nabla_{\boldsymbol{H}_n^{({\cal B})}} L$ 
and sends it to client $n$. Finally, each client $n$ further computes the gradient with respect to its local model, and updates the local model with learning rate $\eta_n$, i.e., $\boldsymbol{W}_n =\boldsymbol{W}_n - \eta_n\nabla_{\vct{H}_n^{({\cal B})}} L \cdot \nabla_{\boldsymbol{W}_n} \boldsymbol{H}_n^{({\cal B})}$.

\subsection{Straggler and privacy challenges}

\noindent {\bf Challenge 1: Performance degradation from stragglers.} Straggler problem is commonly observed in FL systems, due to heterogeneous computation and communication resources across clients, and can be even worse for VFL systems where heterogeneity also exists for local model architecture and data features. 
To understand the effect of stragglers on model performance, we carry out experiments on the FashionMNIST dataset~\cite{xiao2017/online}
in split VFL setting, where 16 clients evenly hold parts of each training image. We select 60\% clients as stragglers to add an additional exponential delay when submitting their embeddings. We compare three strategies to handle stragglers: 1) Wait for all stragglers (Wait); 2) Ignore stragglers (Ignore); and 3) VAFL with asynchronous model updates~\cite{DBLP:journals/corr/abs-2007-06081}. 
Three methods, including concatenation, element-wise average and element-wise maximum, are utilized for embedding aggregation. As shown in Figure \ref{fig:pre}, for all aggregation methods and strategies, presence of stragglers leads to convergence slowdown and accuracy degradation. 

    
\noindent {\bf Challenge 2: Data/model leakage.} The embedding from a client contains information about its private data and local model parameters. It has been shown in~\cite{luo2021feature,erdogan2021unsplit} that through inference attacks, a curious server can reconstruct a victim client's private input features and local model, from its uploaded embedding.

\noindent {\bf Threat model.} We consider an \emph{honest-but-curious} threat model, which is widely adopted to study the privacy vulnerabilities of FL systems. All parties in the system will faithfully follow the specified learning protocol. The curious server attempts to infer private data and local model of a victim client from its uploaded computation results. A subset of curious clients may collude to infer the private data and local models of the other victim clients. 

The goal of this work is to tackle the above challenges, via developing a synchronous split VFL framework whose model training is resilient to stragglers, and private against passively inferring clients' local data and model parameters.




%% file: preliminary.tex
\section{Preliminaries}

\noindent {\bf Embedding averaging.} We adopt the element-wise average as the aggregation method. That is, $ \vct{H}^{(\cal B)}  = \frac{1}{N} \sum_{n=1}^N \vct{H}^{({\cal B})}_n$.
The reason for this choice is two-folded: 1) As shown in Figure~\ref{fig:pre}, compared with concatenation, which is the best performing aggregation method, element-wise average achieves comparable performance when there is no straggler; 
2) For element-wise average, the server does not necessarily need to know \emph{individual} client embeddings to compute their summation, hence potentially allowing a higher level of privacy protection. 
To implement this embedding averaging, we require the same dimension for the embeddings from all clients, i.e., $h_1=\cdots=h_N=h$.

\noindent {\bf Lagrange coded computing.} Proposed in~\cite{yu2019lagrange}, Lagrange coded computing (LCC) is a cryptographic primitive for sharing multiple secrets. 
Given a privacy parameter $T$, LCC guarantees information-theoretic privacy against up to $T$ colluding shares. LCC supports homomorphic evaluation of arbitrary polynomials on the shares. The decryption is accomplished through polynomial interpolation, which is resilient to loss of decryption shares up to a certain threshold.

\noindent {\bf Polynomial networks.} As one of our main goals is to provide data and model privacy for split VFL, which requires utilizing secure computation primitives like LCC, we adopt polynomial network (PN) as the architecture of the client models. Proposed in~\cite{livni2014computational}, a PN uses quadratic function as the activation function, and outputs a polynomial function of the input. 
For instance, the output $y\in \mathbb{R}$ of a $2$-layer PN with $r$ neurons in the hidden layer, for some input $\vct{x} \in \mathbb{R}^d$, is computed as $y = b+ \vct{w}_0^{\top}\vct{x} + \sum_{i = 1}^r\alpha_i(\vct{w}_i^{\top}\vct{x})^2$,
where $\vct{w}_i \in \mathbb{R}^d$ are network parameters. 
Compared with standard architectures like MLP and CNN with non-linear activation functions, PN is natively compatible with homomorphic evaluations on secret shares, and at the same time exhibited superior performance~\cite{DBLP:journals/corr/abs-2106-13834}. Here we consider a simplified architecture such that for a PN with $D$ layers, the output embedding $\vct{h}\in \mathbb{R}^h$ is produced from an input $\vct{x} \in \mathbb{R}^d$ as $
\vct{h} =\sum_{i=1}^{D} (\vct{x}^i \vct{W}^{i} + \vct{b}^{i})$,
where $\vct{x}^{i}$ is the $i$th power of the input computed element-wise, and  $\vct{W}^{i} \in \mathbb{R}^{d \times h}$ and $\vct{b}^{i} \in \mathbb{R}^{h}$ are the weight matrix and bias vector for the $i$th layer. 

\begin{table}[h!t]
\vspace{-5mm}
\centering
\caption{Test accuracies of different client network architectures.
\vspace{1mm}
}
    \begin{tabular}{c | c c c c}
    \hline
     \# of layers & MLP & CNN & PN \\
         \hline
          1 & 88.35\%  & 90.48\% & 88.19\% \\
       
       2&88.60\%  &90.79\% & 88.31\% \\
      
         3 & 88.70\%&91.53\%  & 88.49\%\\
         \hline
    \end{tabular}
    \label{tab:client_network}
    \vspace{-3mm}
\end{table}

In a split VFL system, a PN with $D_n$ layers at client $n$ consists of $D_n$ weight matrices $\vct{W}_n = (\vct{W}_n^1,\ldots,\vct{W}_n^{D_n})$, where $\vct{W}_n^{i}\in \mathbb{R}^{d_n \times h}$. For an input data partition $\vct{X}_n^{({\cal B})}$ of batch ${\cal B}$, the output embeddings are computed as 
\begin{equation}
    \vct{H}_n^{({\cal B})} = g_n(\vct{X}_n^{({\cal B})},\vct{W}_n) = \sum_{i = 1}^{D_n} \vct{X}_n^{i,({\cal B})} \vct{W}_n^{i}, 
    \footnote{The bias vectors are absorbed into the weight matrices with a $1$ appended to each data sample. }
\end{equation}
where $\vct{X}_n^{i,({\cal B})}$ is a matrix whose elements are $i$th power of the corresponding elements in $\vct{X}_n^{({\cal B})}$.



To verify the effectiveness of using PN in split VFL, we train image classifiers on FashionMNIST 
over $4$ clients.  
The server holds a VGG13 network~\cite{simonyan2014very}; three different network architectures, including MLP, CNN, and PN, are respectively employed at the clients. As shown in Table~\ref{tab:client_network}, PN achieves comparable performance with CNN. which has the highest accuracies. 

%% file: protocol.tex
\section{Protocol Description}

\subsection{Overview}
We develop a synchronous split VFL framework \scheme, which simultaneously addresses the straggler and privacy leakage challenges. In \scheme, each client secret shares its training data across the network using LCC before training starts.
In each training round, each client first secret shares its current local model; then, utilizing the algebraic structures of the shares and the underlying PN computation, each client performs homomorphic evaluations on coded data and models, and sends computation results to the server. The summation of embeddings can be reconstructed losslessly at the server, in spite of missing results from a threshold number of stragglers. We give a full description of \scheme in Algorithm~\ref{alg:FedVS}.

\begin{algorithm}
\caption{The \scheme protocol} \label{alg:FedVS}
\textbf{Input:} $K$ (partition parameter), $T$ (privacy parameter) 
\begin{algorithmic}[1]
\STATE // \emph{Data preparation phase}
\FOR {each client $n= 1,2,\ldots,N$ \textbf{in parallel}}
    \STATE 
    $\widehat{\vct{X}}_n 
 \gets (\vct{X}_n^1, \ldots, \vct{X}_n^{D_n})$ // Raises data to the degree of local PN
     \STATE $\overline{\vct{X}}_n  \gets$ Quantization on $\widehat{\vct{X}}_n$ 
    \STATE $\overline{\vct{X}}_{n,1},\ldots,\overline{\vct{X}}_{n,K} \gets$ Horizontally partitions $\overline{\vct{X}}_n$ into $K$ segments
    \STATE $\vct{Z}_{n,K+1},\ldots,\vct{Z}_{n,K+T}\gets$ Sample random masks
    \STATE $\{\widetilde{\vct{X}}_{n,n'}\}_{n' \in [N]}\gets$ Evaluating (\ref{eq:data_encoding}) at $\alpha_1,\ldots,\alpha_N$ // Data secret shares
   \STATE Sends data share $\widetilde{\vct{X}}_{n,n'}$ to client $n' \in [N]\backslash \{n\}$
   \STATE Receives data share $\widetilde{\vct{X}}_{n',n}$  from user $n' \in [N]\backslash \{n\}$
\ENDFOR

\item[]
\STATE // \emph{Training phase}
\FOR{Round $1,2,\ldots$}
\STATE // \emph{Model secret sharing}\\
\FOR {each client $n= 1,2,\ldots,N$ \textbf{in parallel}}
\STATE $\overline{\vct{W}}_n \gets$ Quantization on $\vct{W}_n$
\STATE $\vct{V}_{n,K+1},\ldots,\vct{V}_{n,K+T}\gets$ Sample random masks
    \STATE $\{\widetilde{\vct{W}}_{n,n'}\}_{n' \in [N]}\gets$ Evaluating (\ref{eq:model_encode}) at $\alpha_1,\ldots,\alpha_N$ // Model secret shares
   \STATE Sends model share $\widetilde{\vct{W}}_{n,n'}$ to client $n' \in [N]\backslash \{n\}$
   \STATE Receives model share $\widetilde{\vct{W}}_{n',n}$  from user $n' \in [N]\backslash \{n\}$
\ENDFOR

\STATE // \emph{Homomorphic embedding evaluation}\\
\FOR {each client $n= 1,2,\ldots,N$ \textbf{in parallel}}
   \STATE For a sample batch ${\cal B}$, computes coded embedding $\widetilde{\vct{H}}^{({\cal B})}_{n}$ as in (\ref{eq:evaluation}) and sends it to server
\ENDFOR
\STATE // \emph{Server model update} \\
\STATE \textbf{Server executes:}
\STATE Receives coded embeddings from non-straggling clients ${\cal U} \subset [N]$
\STATE Interpolates embedding summation polynomial $\psi(x)$ in (\ref{eq:aggregation_poly}) from $\{\widetilde{\vct{H}}^{({\cal B})}_{n}: n \in {\cal U}\}$
\STATE Recovers embedding summation $\overline{\vct{H}}^{({\cal B})}$ by evaluating $\psi(x)$ at $\beta_1,\ldots,\beta_K$
\STATE $\vct{H}^{({\cal B})} \gets$ Dequantization on $\overline{\vct{H}}^{({\cal B})}$ // Recovers average embedding over all clients (including stragglers)\\
\STATE Back-propogates to update central model $\vct{W}_0$, and broadcasts $\nabla_{\vct{H}^{({\cal B})}} L$ to all clients
\STATE // \emph{Client model update} \\
\FOR {each client $n= 1,2,\ldots,N$ \textbf{in parallel}}
\STATE Obtains $\nabla_{\boldsymbol{W}_n} L \gets \nabla_{\vct{H}_n^{({\cal B})}} L \cdot \nabla_{\vct{W}_n} \vct{H}_n^{({\cal B})}$, and  updates local model $\vct{W}_n$ 
\ENDFOR
\ENDFOR
\end{algorithmic}
\end{algorithm}

\subsection{Data preparation}
Before training starts, a data preparation step takes place among the clients.

\noindent {\bf Pre-processing and quantization.} 
Each client $n$ pre-processes its input $\vct{X}_n$ to obtain $\widehat{\vct{X}}_n = (\vct{X}_n^1, \ldots, \vct{X}_n^{D_n})$, where $\vct{X}_n^i$ is computed via raising $\vct{X}_n$ to the $i$th power element-wise. 
Then, the client quantizes $\widehat{\vct{X}}_n$ onto a finite field $\mathbb{F}_p$, for some sufficiently large prime $p$.
Specifically, for some scaling factor $l_x$, rounding operator $\operatorname{Round}(x)= \begin{cases}\left\lfloor x\right\rfloor, & \textup{ if } \;  x -\left\lfloor x\right\rfloor<0.5 \\ \left\lfloor x\right\rfloor+1, & \text { otherwise }\end{cases}$, and shift operator $\phi(x)= \begin{cases}x, & \text { if } x \geq 0 \\ p+x, & \text { if } x<0\end{cases}$, client $n$ obtains its quantized data $\overline{\vct{X}}_n = \phi (\operatorname{Round}(2^{l_x}\cdot \widehat{\vct{X}}_n))$, applied element-wise.

\noindent {\bf Private data sharing.} The clients secret share their quantized local data with other clients using LCC with \emph{partition parameter} $K$ and \emph{privacy parameter} $T$. Specifically, for each $n \in [N]$, client $n$ horizontally partitions $\overline{\vct{X}}_n=(\overline{\vct{X}}_n^1, \ldots, \overline{\vct{X}}_n^{D_n})$ into $K$ segments $\overline{\vct{X}}_{n,1},\ldots,\overline{\vct{X}}_{n,K}$, and then samples independently $T$ masks $\vct{Z}_{n,K+1},\ldots,\vct{Z}_{n,K+T}$ uniformly at random. For a set of distinct parameters $\{\beta_1,\ldots,\beta_{K+T}\}$ from $\mathbb{F}_p$ that are agreed among all clients and the server, using Lagrange interpolation, client $n$ obtains the following polynomial. 
\begin{equation}
\label{eq:data_encoding}
\begin{aligned}
\vct{F}_n(x)= \sum\limits_{k=1}^{K}\overline{\vct{X}}_{n,k} \cdot \prod_{\ell \in[K+T]\backslash\{k\}}\frac{x-\beta_{\ell}}{\beta_{k}-\beta_{\ell}}  \\
+\sum\limits_{k=K+1}^{K+T}\vct{Z}_{n,k} \cdot \prod_{\ell \in[K+T]\backslash\{k\}}\frac{x-\beta_{\ell}}{\beta_{k}-\beta_{\ell}}.
\end{aligned}
\end{equation}
Here we note that $F_{n}(\beta_k) = \overline{\vct{X}}_{n,k}$, for all $k \in [K]$. 

For another set of public parameters $\{\alpha_1,\ldots,\alpha_N\}$ that are pair-wise distinct and $\{\beta_1,\ldots,\beta_{K+T}\} \cap \{\alpha_1,\ldots,\alpha_N\} = \emptyset$, client $n$ computes $\widetilde{\vct{X}}_{n,n'} = \vct{F}_n(\alpha_{n'})$, for all $n' \in [N]$, and sends it to client $n'$. Note that the size of a secret share is $\frac{1}{K}$ of the size of the original data. Data partitioning in LCC helps to reduce the communication cost for secret sharing, and the complexity of subsequent computations on secret shares. By the end of the data sharing phase, each client $n'$ has locally the secret shares $\widetilde{\vct{X}}_{n'} = (\widetilde{\vct{X}}_{1,n'}, \ldots, \widetilde{\vct{X}}_{N,n'})$ from all $N$ clients.

\subsection{Training operations}
\label{sec:training}{\bf Model quantization and secret sharing.} A training round starts with each client $n$ quantizing and secret sharing its current model parameters $\vct{W}_n$. Firstly, For some scaling factor $l_w$, client $n$ quantizes its model parameters to obtain
\begin{equation}\label{eq:model_quantize}
    \overline{\vct{W}}_n \!=\! ( \overline{\vct{W}}_n^1,\ldots, \overline{\vct{W}}_n^{D_n}) \!=\! \phi (\operatorname{Round}_{stoc}(2^{l_w} \cdot \vct{W}_n)).
\end{equation}
Here $ \operatorname{Round}_{stoc}(x)\!=\! \begin{cases}\left\lfloor x\right\rfloor & \text{with prob.  } 1 - (x -\left\lfloor x\right\rfloor)\\ \left\lfloor x\right\rfloor+1 & \text{with prob. } x -  \left\lfloor x\right\rfloor \end{cases}$ is an unbiased stochastic rounding operator, i.e., $\mathbb{E}[\operatorname{Round}_{stoc}(x)] =x$.



Then, client $n$ samples uniformly at random $T$ noise terms $\vct{V}_{n,K+1},\ldots,\vct{V}_{n,K+T}$, and constructs the following Lagrange polynomial.
\begin{equation}\label{eq:model_encode}
\begin{aligned}
\vct{G}_n(x)=\sum\limits_{k=1}^{K}\overline{\vct{W}}_{n} \cdot \prod_{\ell \in[K+T]\backslash\{k\}}\frac{x-\beta_{\ell}}{\beta_{k}-\beta_{\ell}} \\+\sum\limits_{k=K+1}^{K+T}\vct{V}_{n,k} \cdot \prod_{\ell \in[K+T]\backslash\{k\}}\frac{x-\beta_{\ell}}{\beta_{k}-\beta_{\ell}}. 
\end{aligned}
\end{equation}
For each $n' \in [N]$, client $n$ computes a secret share of it model $\widetilde{\vct{W}}_{n,n'} = \vct{G}_n(\alpha_{n'})$, and sends it to client $n'$.\footnote{WLOG, we assume that all clients successfully share their models with all other clients. In a more general scenario where each client may not be able to communicate with every other client, we can consider a subset $\mathcal{S}$ of clients who have successfully shared their models with a subset $\mathcal{R}$ of clients, and the proposed FedVS protocol can be used to compute the aggregated embedding $\sum_{n \in {\cal S}} \vct{H}^{({\cal B})}_n$ from the uploaded results of clients in ${\cal R}$.}


\noindent {\bf Homormophic evaluation and embedding decryption.} For a batch ${\cal B} \subseteq [\frac{M}{K}]$ of coded training samples, the clients start forward propagation by homomorphic embedding evaluation. Specifically, for each $n' \in [N]$, client $n'$ takes the coded data $\widetilde{\vct{X}}^{({\cal B})}_{1,n'},\ldots,\widetilde{\vct{X}}^{({\cal B})}_{N,n'}$, with $\widetilde{\vct{X}}^{({\cal B})}_{n,n'} = (\widetilde{\vct{X}}^{1, ({\cal B})}_{n,n'},\ldots,\widetilde{\vct{X}}^{D_n,({\cal B})}_{n,n'})$ for all $n \in [N]$, and the coded models $ \widetilde{\vct{W}}_{1,n'},\ldots,\widetilde{\vct{W}}_{N,n'}$, with $\widetilde{\vct{W}}_{n,n'} = (\widetilde{\vct{W}}^1_{n,n'} \ldots, \widetilde{\vct{W}}^{D_n}_{n,n'})$ for all $n \in [N]$, computes its output
\begin{align}\label{eq:evaluation}
    \widetilde{\vct{H}}^{({\cal B})}_{n'} \!\!=\!\! \sum_{n=1}^N \! g_{n}(\widetilde{\vct{X}}^{({\cal B})}_{n,n'}, \widetilde{\vct{W}}_{n,n'}) \!=\!\!\!\sum_{n=1}^N \!\sum_{i = 1}^{D_n} \!\widetilde{\vct{X}}_{n,n'}^{i,({\cal B})} \widetilde{\vct{W}}_{n,n'}^i,
\end{align}
and sends $\widetilde{\vct{H}}^{({\cal B})}_{n'}$ to the server. During this process, some clients become stragglers, and server only waits to receive results from a subset ${\cal U} \subset [N]$ of non-straggling clients. 

It is easy to see that for the polynomial $\vct{F}^{({\cal B})}_n(x) = (\vct{F}^{1,({\cal B})}_n(x) \ldots, \vct{F}^{D_n,({\cal B})}_n(x))$ corresponding to data batch ${\cal B}$, and the model polynomial $\vct{G}_n(x) = (\vct{G}^1_n(x)\ldots,\vct{G}^{D_n}_n(x))$, $\widetilde{\vct{H}}^{({\cal B})}_{n'}$ can be viewed as the evaluation of the following composite polynomial at point $x = \alpha_{n'}$.
\begin{align}
\label{eq:aggregation_poly}
    \psi(x) = \sum_{n=1}^N \sum_{i=1}^{D_n} \vct{F}^{i,({\cal B})}_n(x) \vct{G}^{i}_n(x).
\end{align}
The server interpolates $\psi(x)$ from the received results $(\widetilde{\vct{H}}^{({\cal B})}_{n'})_{n' \in {\cal U}}$, and evaluates it at $\beta_1,\ldots,\beta_K$ to recover the summation of the embedding segments $\sum_{n=1}^N\overline{\vct{H}}_{n,1}^{({\cal B})},\ldots, \sum_{n=1}^N\overline{\vct{H}}_{n,K}^{({\cal B})}$, where $\sum_{n=1}^N\overline{\vct{H}}_{n,k}^{({\cal B})}= \psi(\beta_k) = \sum_{n=1}^N \sum_{i=1}^{D_n} \overline{\vct{X}}^{i,({\cal B})}_{n,k} \overline{\vct{W}}^{i}_n$. The server horizontally stacks these summed segments to obtain the summation 
 $\overline{\vct{H}}^{({\cal B})}$ of local embeddings. Note that the overall batch size of $\overline{\vct{H}}^{({\cal B})}$
is $K|{\cal B}|$.


\noindent \textbf{Dequantization.} The server maps $\overline{\vct{H}}^{({\cal B})}$ back to the real domain 
to obtain an approximation of the average embedding $\vct{H}^{({\cal B})}$ via applying the following dequantization function $\varphi: \mathbb{F}_p \rightarrow \mathbb{R}$ element-wise on $\overline{\vct{H}}^{({\cal B})}$.
\begin{equation}
\varphi(x) \!=\!\! \begin{cases} \frac{1}{N} \cdot 2^{-(l_x+l_w)} \cdot x, & \text{ if  } 0 \leq x < \frac{p-1}{2} \\
\frac{1}{N} \cdot 2^{-(l_x+l_w)} \cdot (x - p), & \text{ if  } \frac{p-1}{2} \leq x < p \end{cases}.
\end{equation}

Next, server continues forward-backward propagation to update the central model $\vct{W}_0$. The server also computes $\nabla_{\vct{H}^{({\cal B})}} L$, and broadcasts it to all clients. 
With $\nabla_{\vct{H}_n^{({\cal B})}} L =\frac{1}{N}\nabla_{\vct{H}^{({\cal B})}} L $, client $n$ computes the gradient $\nabla_{\boldsymbol{W}_n} L=\nabla_{\vct{H}_n^{({\cal B})}} L \cdot \nabla_{\vct{W}_n} \vct{H}_n^{({\cal B})}$, and updates its local model $\vct{W}_n$. 

%% file: analysis.tex
\section{Theoretical Analyses}

\subsection{Straggler resilience and privacy analysis}

\begin{theorem}[Straggler resilience]\label{thm:straggler_resilience}
The summation of local embeddings of all clients, i.e, $\overline{ \vct{H}}^{({\cal B})} =\sum_{n=1}^N\overline{\vct{H}}_n^{({\cal B})}$, can be exactly recovered at the server, in the presence of up to $N-2(K+T-1)-1$ straggling clients. 
\end{theorem} 

\begin{proof}
The server can exactly reconstruct $\psi(x)$, and hence the summation of local embeddings $\overline{ \vct{H}}^{({\cal B})}$, from the computation results of the non-straggling clients $(\widetilde{\vct{H}}^{({\cal B})}_{n})_{n \in {\cal U}}$, if $|{\cal U}| \geq \textup{degree}(\psi(x))+1 = 2(K+T-1)+1$. Hence, the embedding aggregation process can tolerate up to $N-2(K+T-1)-1$ stragglers.
\end{proof}

\begin{theorem}[Privacy against colluding clients]\label{thm:privacy-clients}
Any subset of up to $T$ colluding clients learn nothing about the local data and models of the other clients. More precisely, for any ${\cal T} \subset [N]$ with $|{\cal T}| \leq T$, the mutual information $I\left((\widetilde{\vct{X}}_n, \widetilde{\vct{W}}_n)_{n \in {\cal T}};(\overline{\vct{X}}_n, \overline{\vct{W}}_n)_{n \in [N]\backslash {\cal T}}\right)$ equals zero.
\end{theorem}

\begin{proof}
As the local data and models are secret shared using LCC, their privacy against $T$ colluding clients follows the $T$-privacy guarantee of LCC construction ( Theorem~1 in~\cite{yu2019lagrange}). For completeness, we give a detailed proof in Appendix~\ref{sec:privacy_proof}.
\end{proof}

\begin{theorem}[Privacy against curious server]\label{thm:privacy-server}
For each $n \in [N]$, the server learns nothing about the private data and the local model of client $n$, from its uploaded computation result. That is, the mutual information $I\left(\widetilde{\vct{H}}_n^{({\cal B})};(\overline{\vct{X}}_n, \overline{\vct{W}}_n)\right)$ equals zero.
\end{theorem}

\begin{proof}
We know from the privacy guarantee of LCC that the secret shares of input data and model parameters at client $n$, i.e., $(\widetilde{\vct{X}}_n, \widetilde{\vct{W}}_n)$, reveal no information about its private data and model $(\overline{\vct{X}}_n, \overline{\vct{W}}_n)$. Moreover, as the output $\widetilde{\vct{H}}_n^{({\cal B})}$ of client $n$ is computed from $(\widetilde{\vct{X}}_n, \widetilde{\vct{W}}_n)$, i.e., $(\overline{\vct{X}}_n, \overline{\vct{W}}_n) \rightarrow (\widetilde{\vct{X}}_n, \widetilde{\vct{W}}_n) \rightarrow \widetilde{\vct{H}}_n^{({\cal B})}$ forms a Markov chain, and we have $I\left(\widetilde{\vct{H}}_n^{({\cal B})};(\overline{\vct{X}}_n, \overline{\vct{W}}_n)\right) \leq I\left((\widetilde{\vct{X}}_n, \widetilde{\vct{W}}_n);(\overline{\vct{X}}_n, \overline{\vct{W}}_n)\right) = 0$ by data processing inequality. 
\end{proof}
Theorem~\ref{thm:privacy-server} implies that in \scheme, local data and model of an individual client is \emph{perfectly} secure against the server, which completely mitigates 
\emph{any} privacy leakage from 
data inference and model stealing attacks on a client's output.

\subsection{Convergence analysis}


Since the rounding operation can be performed on both training and test data, \scheme can be considered to optimize the model parameters on the rounded data, which is denoted as $(\vct{x}^{\prime(m)},\vct{y}^{(m)}), m\in [M]$. That is, we consider the following optimization problem for $\vct{W} = (\vct{W}_n)_{n=0}^N$.
\begin{equation*}
\begin{split}
       \min_{\vct{W}} \! F(\vct{W}) 
       \!\triangleq\! \frac{1}{M}\!\! \sum_{m=1}^M \!\! \ell (\vct{W}; \!(\vct{x}^{\prime(m)},\vct{y}^{(m)})) \!=\!\!  \frac{1}{M}\!\! \sum_{m = 1}^{M} \!f_m(\vct{W}).
\end{split}
\end{equation*}

In round $r$ of \scheme, for a sampled data batch ${\cal B}$, the server and the clients update their models as $\vct{W}_n^{r+1} = \vct{W}_n^{r} - \eta_n\nabla_n F_{\cal B}(\widehat{\vct{W}}^r)$, $\forall n\in \{0,1,\ldots,N\}$, where $F_{\cal B}(\cdot) = \frac
{1}{|\cal B|}\sum_{m\in \cal B}f_m(\cdot)$. Here we have $\widehat{\vct{W}}_0^r = \vct{W}_0^r$, as the central model is not rounded during forward propogation; for each client $n$, $\widehat{\vct{W}}^r_n \!=\! Q_{stoc}(\vct{W}^r_n) \!=\! 2^{-l_w} \cdot \operatorname{Round}_{stoc}(2^{l_w}\cdot\vct{W}^r_n)$.

We first make the following assumptions 
to facilitate our convergence analysis.

\textbf{Assumption 1 (Variance-bounded stochastic rounding):} There exists a constant $\gamma > 0$ such that $\forall z \in \mathbb{R}$, the operator 
$Q_{stoc}(.)$ satisfies $\mathbb{E}\left[\|Q_{stoc}(z) - z\|^2\right]\leq \gamma^2z^2 $.

\textbf{Assumption 2 (Lipschitz Smoothness):} For any input $\vct{u}, \vct{v}$, there exists a constant $L > 0$, such that for all $m\in[M]$, the function $f_m$ satisfies $\forall n \in \{0,1,\ldots,N\}$,
$\|\nabla_n f_m(\vct{u}) - \nabla_n f_m(\vct{v})\| \leq L \|\vct{u} - \vct{v}\|$.

\textbf{Assumption 3 (Global minimum existance):} There exists a globally optimal collection of model parameters $\vct{W}^{*}$, such that $F(\vct{W}) \geq F(\vct{W}^{*}) > -\infty$, for all $\vct{W}$.


\textbf{Assumption 4 (Bounded model parameters):} The norm of the collection of all model parameters $\|\vct{W}\|$ is upper bounded by some constant $\sigma$. 



We give the convergence result of FedVS in the following theorem, whose proof can be found in Appendix~\ref{sec:convergence_proof}.
\begin{theorem}[Convergence of FedVS]\label{thm4}
Under Assumption 1-4, when the learning rate $\eta_n =\frac{3}{4L} \frac{1}{\sqrt{R}}, \forall n\in \{0,1,\ldots,N\}$, after $R$ rounds of \scheme, with probability at least $1-\delta$ we have:
\begin{equation*}
\begin{split}
    \frac{1}{R}\sum_{r=0}^{R-1} \! \mathbb{E} \big( \! \sum_{n=0}^N&\|\nabla_n F(\vct{W}^r)\|^2 \big) \! \leq \! \frac{16L}{9\sqrt{R}} (F(\vct{W}^{0}) \!-\! F(\vct{W}^{*})) \\
&+\frac{\sum_{n=0}^N(2L^2\gamma^2\sigma^2 + 2V_n) }{\sqrt{R}} = \mathcal{O} \left(\frac{1}{\sqrt{R}}\right),
    \end{split}
\end{equation*}
where $V_n \!\!=\!\! \frac{32 L^2(\log(2p_n/\delta)+\frac{1}{4})}{|\cal B|}$, and $p_n$ is dimension of $\vct{W}_n$.
\end{theorem}

\subsection{Complexity analysis}

{\bf Computation and communication costs for data sharing.} Before training starts, each client secret shares its local data 
using LCC. Given that evaluating a polynomial of degree $K+T-1$ at $N$ points can be done using $\mathcal{O}(N \log^2 N)$ operations in $\mathbb{F}_p$~\cite{von2013modern}, the computation load at client $n$ to generate $N$ shares is $\mathcal{O}(\frac{M d_n D_n}{K}N\log^2N)$. 
The communication cost for client $n$ to secret share its data is $\mathcal{O}(\frac{M d_n D_n N}{K})$. We note that these computation and communication overheads occur once before the training starts, and become less relevant as the number of training rounds increases.  


{\bf Computation and communication costs for a training round.} In each training round, each client $n$ first needs to secret share its local model, which takes a computation load of $\mathcal{O}(d_n h D_n N\log^2N)$ and a communication load of $\mathcal{O}(d_n h D_n N)$. Next, for the sampled data batch ${\cal B}$ of size $|{\cal B}| \leq \frac{M}{K}$, client $n$ performs embedding computation as in~(\ref{eq:evaluation}) with a computation load of $\mathcal{O}(|{\cal B}|h \sum_{i=1}^N d_iD_i)$, and sends the computed results to the server with a communication load of $\mathcal{O}(|{\cal B}|h)$. Note that while according to Theorem~\ref{thm:straggler_resilience} a smaller partition paramter $K$ allows to tolerate more stragglers, the load of embedding computation 
is also higher. 
We stress that in \scheme, the loads of computing and communicating (coded) embeddings are identical across all clients, further alleviating the straggler effect caused by imbalanced data and model dimensions.

Server decodes the embedding aggregation from results of $R\!=\!2(K\!+T\!-1)+1$ non-straggling clients, via interpolating $\psi(x)$ in (\ref{eq:aggregation_poly}) with a computation cost of $\mathcal{O}(|{\cal B}|h R\log^2 R)$.

%% file: experiment.tex
\section{Experimental Evaluations}

We carry out split VFL experiments on three types of six real-world datasets, and compare the performance of \scheme in straggler mitigation and privacy protection with four baselines.
All experiments are performed on a single machine using four NVIDIA GeForce RTX 3090 GPUs. 


\begin{figure*}[ht]

\minipage{0.33\textwidth}
  \centering
   {\quad CV}
\endminipage\hfill
\minipage{0.33\textwidth}
  \centering
 {\quad Multi-view}
\endminipage\hfill
\minipage{0.33\textwidth}%
  \centering
 {\quad Tabular}
  \endminipage
\\
\vspace{-4mm}

\minipage{0.33\textwidth}
  \centering\includegraphics[width=\columnwidth]{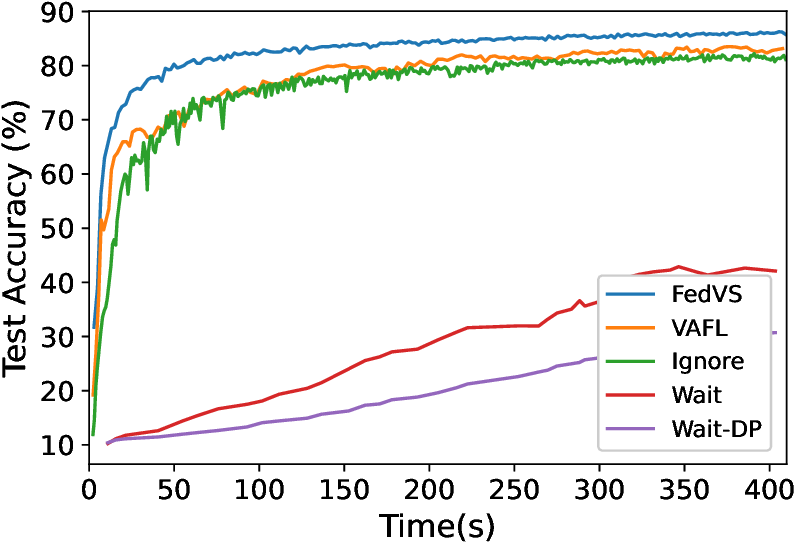}
  \subfigure{\small{\quad (a) FashionMNIST}}%
\endminipage\hfill
\minipage{0.33\textwidth}
  \centering
  \includegraphics[width=\columnwidth]{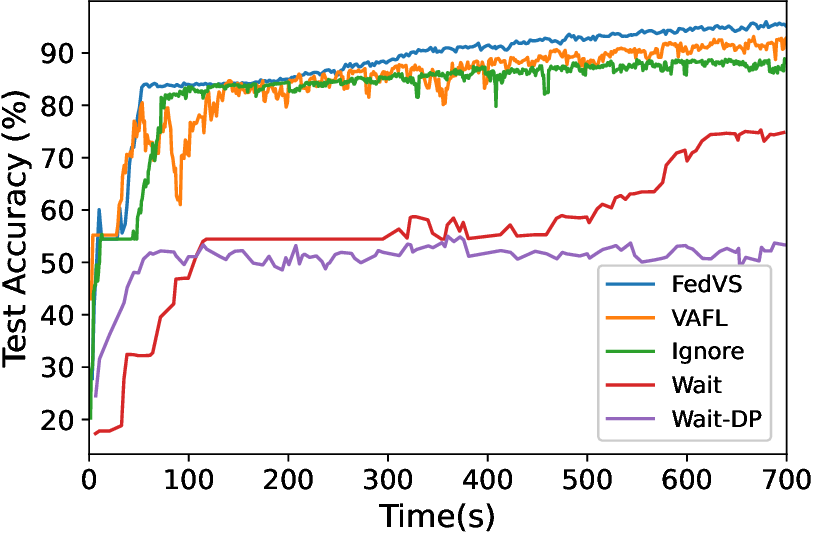}
  \subfigure{\small{\quad (c) Caltech-7}}
\endminipage\hfill
\minipage{0.33\textwidth}%
  \centering
  \includegraphics[width=\linewidth]{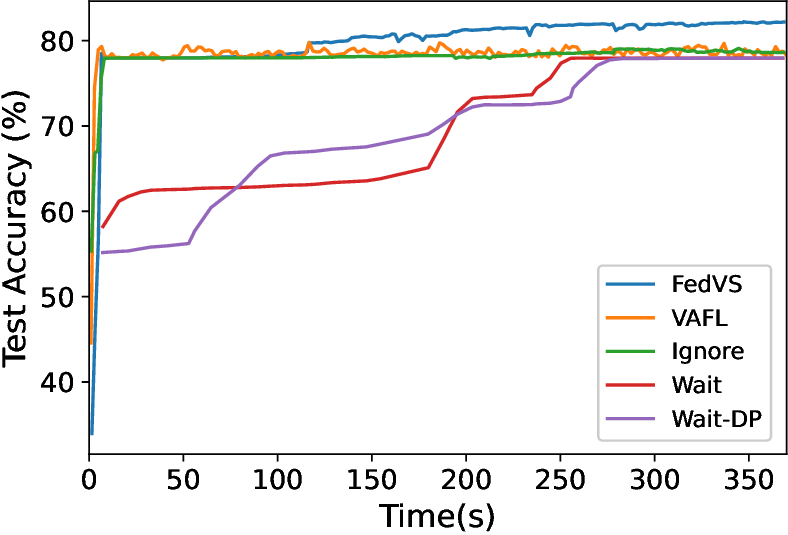}
  \subfigure{\small{\quad (e) Credit card}}
  \endminipage
\quad
\minipage{0.33\textwidth}%
  \centering
  \includegraphics[width=\linewidth]{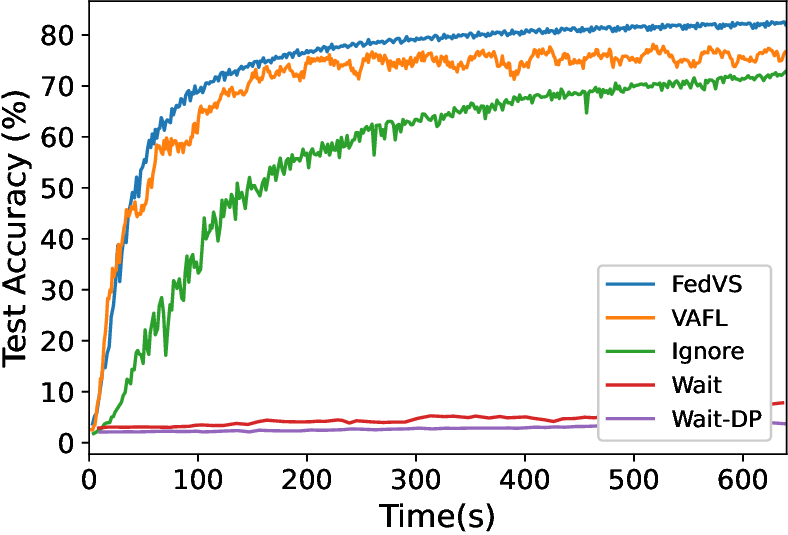}

\subfigure{ \small{(b) EMNIST}}
  \endminipage
  \minipage{0.33\textwidth}%
  \centering
  \includegraphics[width=\linewidth]{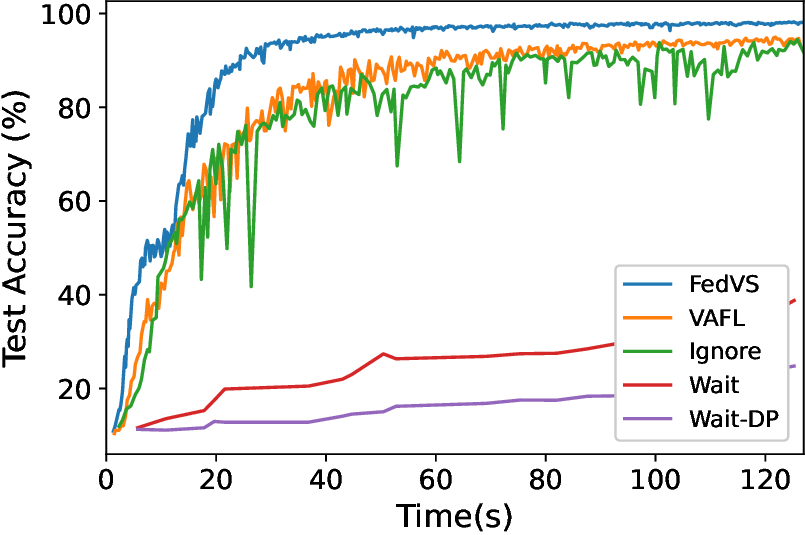}
 \subfigure{\small{\quad (d) HandWritten}}
  \endminipage
  \minipage{0.33\textwidth}%
  \centering
  \includegraphics[width=\linewidth]{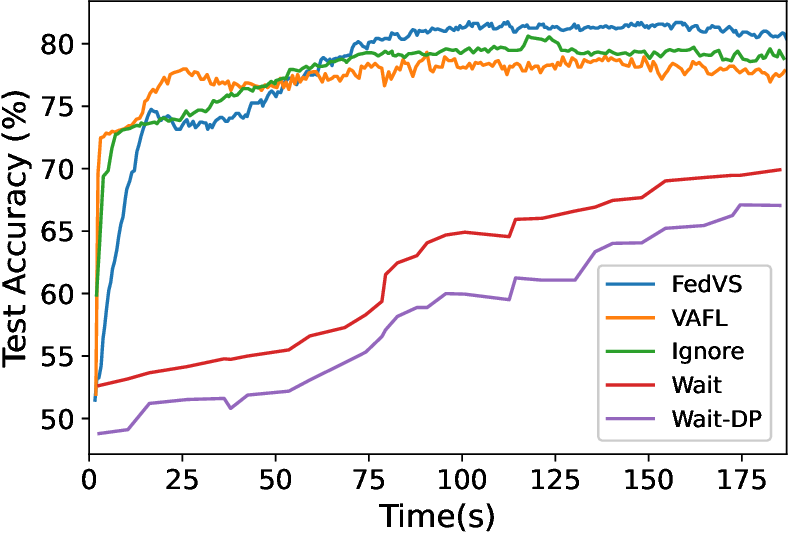}
\subfigure{\small{\qquad (f) Parkinson}}
  \endminipage
\vspace{-3mm}
\caption{Test accuracies using different straggler handling and privacy protection methods on different datasets. 
}
\label{sixdata}
\end{figure*}
\begin{figure*}[ht]
\minipage{0.66\textwidth}
\centering
 \subfigure[Delay dominant.]{
\includegraphics[width=0.4795\linewidth, scale=.4]{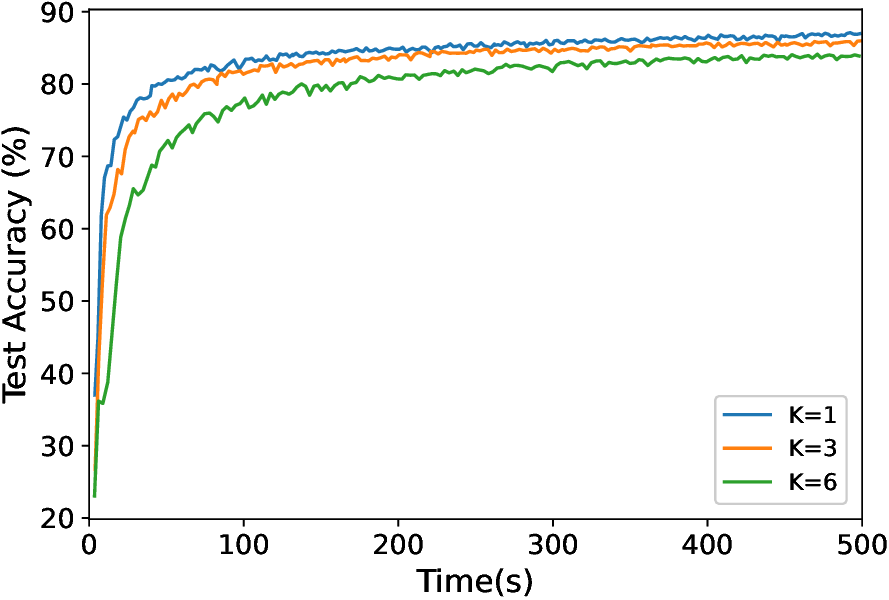}
  }
  \subfigure[Computation dominant.]{
\includegraphics[width=0.4815\linewidth]{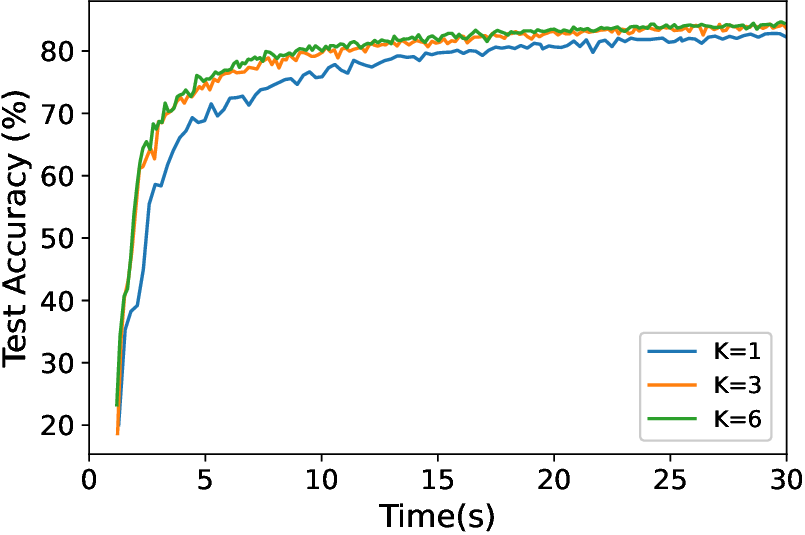}
  }
\caption{Test accuracies of \scheme on FashionMNIST using different $K$.}
\label{Ksele}
\endminipage
\minipage{0.01\textwidth}
\centering
\text{\quad}
\endminipage
  \minipage{0.33\textwidth}%
  \subfigure{
  \includegraphics[width=\linewidth]
  {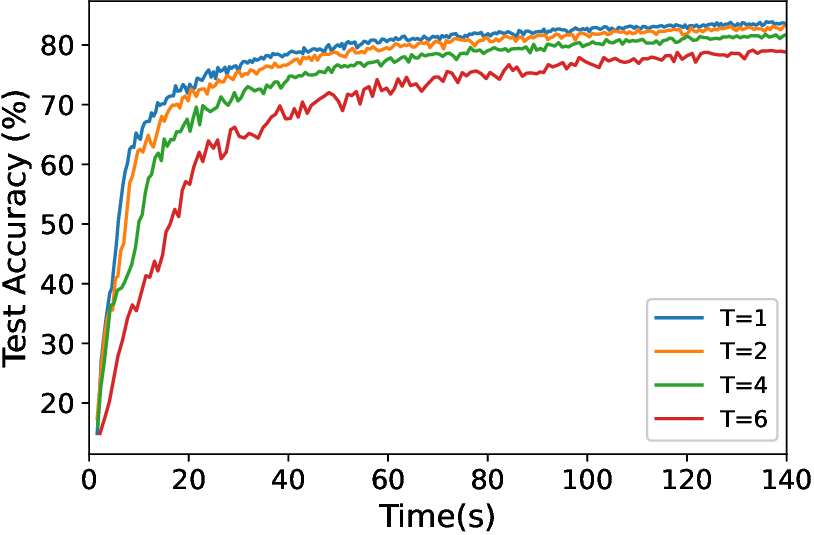}
  }
 \caption{Test accuracies of \scheme on FashionMNIST using different $T$.}
\label{Tsele}
  \endminipage
  \vspace{-2mm}
\end{figure*}

\subsection{Datasets}
We consider three types of data, and select two datasets from each type.
For tabular datasets Parkinson~\cite{sakar2019comparative} and Credit card~\cite{yeh2009comparisons}, and computer vision (CV) datasets EMNIST~\cite{cohen2017emnist} and FashionMNIST~\cite{xiao2017fashion}, we evenly partition the features of each data sample across the clients. 
For the multi-view datasets Handwritten~\cite{Dua:2019} and Caltech-7~\cite{li_andreeto_ranzato_perona_2022}, each client holds one view of each data sample.
We provide descriptions of the datasets, number of clients considered for each dataset, employed model architectures, and training parameters in Appendix~\ref{sec:dataset}.

\subsection{Experiment settings}
\textbf{Baselines.} We consider the following four baseline methods for straggler handling and privacy protection. 1) \emph{Wait}: Server waits for all clients (including stragglers) for embedding aggregation; 2) \emph{Ignore}: Server ignores the stragglers, and proceeds with aggregating embeddings from non-stragglers; 3) \emph{VAFL}~\cite{DBLP:journals/corr/abs-2007-06081}: Server asynchronously receives embeddings and updates model parameters; 4) \emph{Wait-DP}: To utilize differential privacy to protect clients' data and model privacy, as in~\cite{thapa2022splitfed}, a calibrated noise is added to the output (e.g., embedding) of a network layer at each client, and the server waits to aggregate all clients' perturbed embeddings.


{\bf Delay pattern.} We add artificial delays to the clients' computations to simulate the effect of stragglers. Before the clients upload their computed embeddings, 50\% of them add a random delay sampled from an exponential distribution with a mean of 0.1s. The other 50\% are modelled as stragglers, whose delays are sampled from exponential distributions with incremental means, i.e., $2+ \frac{4i}{N}, i\in[\frac{N}{2}]$. 
Besides, the straggler effect in the model sharing phase of \scheme is also simulated by adding an exponential delay at each client, whose mean, according to the analysis of computation costs, is $(\log^{2}N)/|\cal B|$ times of the corresponding delay's mean for  embedding uploading.

\textbf{Parameter settings.} For Wait-DP, we set the privacy budget $\epsilon'$ to 5. For FedVS, 
we optimize the rate of convergence over the partition parameter $K$ for each dataset. The privacy parameter of \scheme is set to $T = 1$. To simulate communication delays, a network bandwidth of 300Mbps, as measured in~\cite{so2022lightsecagg} for AWS EC2 cloud computing environment, is assumed for the server and all clients. Each experiment is repeated 5 times and the average accuracies are reported.


\subsection{Results}
{\bf Comparisons with baselines.} As shown in Figure~\ref{sixdata}, for CV and multi-view datasets, \scheme outperforms all baselines in test accuracy at all times. For tabular datasets, VAFL and Ignore converge quickly at the beginning and are eventually outperformed by \scheme. 
For privacy protection, inserting DP noises in Wait-DP hurts the accuracies for all datasets. In sharp contrast, \scheme protects data and model privacy without performance loss.


{\bf Optimization of partition parameter.} We further explore the optimal choices of the partition parameter $K$ for \scheme under different straggler patterns. Specifically, we consider two delay patterns depending on whether the mean of clients' added delays is greater than the local computation time at a single client. As shown in Figure~\ref{Ksele}(a), when the mean delay is greater than the computation time, stragglers cause major performance bottleneck, and it is preferable to use a smaller $K$ to tolerate more stragglers. On the other hand, when the local computation time dominates the delay caused by stragglers, Figure~\ref{Ksele}(b) indicates that it is optimal to choose a larger $K$ to minimize local computation load.





{\bf Privacy-performance tradeoff.} For a larger privacy parameter $T$, the privacy guarantee of \scheme becomes stronger as it protects data and model privacy from $T$ colluding clients. However, as shown in Figure~\ref{Tsele}, its performance suffers as it tolerates less number of stragglers.

%% file: conclusion.tex
\section{Conclusion}
We propose FedVS, a synchronous split VFL framework that simultaneously addresses the problems of straggling clients  and privacy leakage. 
Through efficient secret sharing of data and model parameters and descryption on the computation shares, \scheme losslessly aggregates embeddings from all clients, in presence of a certain number of stragglers; and simultaneously provides information-theoretic privacy against the curious server and a certain number of colluding clients.
Extensive experiments on various VFL tasks and datasets further demonstrate the superiority of \scheme in straggler mitigation and privacy protection over baseline methods.

\section*{Acknowledgement}
This work is in part supported by the National Nature Science Foundation of China (NSFC) Grant 62106057, Guangzhou Municipal Science and Technology Guangzhou-HKUST(GZ) Joint Project 2023A03J0151 and Project 2023A03J0011, Foshan HKUST Projects FSUST20-FYTRI04B, and Guangdong Provincial Key Lab of Integrated Communication, Sensing and Computation for Ubiquitous Internet of Things.

%% file: appendix.tex
\section*{Appendix}

\section{Proof of Theorem \ref{thm:privacy-clients} }\label{sec:privacy_proof}
Here we prove information-theoretic data privacy against $T$ colluding clients, and the proof for model privacy follows the similar steps.

WLOG, let us consider the first $T$ clients colluding to infer private data $\overline{\vct{X}}_n$ of client $n > T$. We have from (\ref{eq:data_encoding}) that the secret shares of $\overline{\vct{X}}_n$ at the first $T$ clients are 
\begin{equation}
\underbrace{
\begin{pmatrix}
\widetilde{\vct{X}}_{n,1}\\
\widetilde{\vct{X}}_{n,2}\\
\vdots\\
\widetilde{\vct{X}}_{n,T}
\end{pmatrix}
}_{\widetilde{\vct{X}}_{n,T}}
=
\underbrace{
\begin{pmatrix}
a_{1,1}  & \hdots & a_{1,K}\\
a_{2,1}  & \hdots & a_{2,K}\\
\vdots  & \hdots & \vdots\\
a_{T,1}  & \hdots & a_{T,K} 
\end{pmatrix}
}_{\vct{A}_1}
\underbrace{
\begin{pmatrix}
\overline{\vct{X}}_{n,1}\\
\overline{\vct{X}}_{n,2}\\
\vdots \\
\overline{\vct{X}}_{n,K}
\end{pmatrix}
}_{\overline{\vct{X}}_n}
+ 
\underbrace{
\begin{pmatrix}
a_{1,K+1}  & \hdots & a_{1,K+T}\\
a_{2,K+1}  & \hdots & a_{2,K+T}\\
\vdots & \hdots & \vdots\\
a_{T,K+1} & \hdots & a_{T,K+T} 
\end{pmatrix}
}_{\vct{A}_2}
\underbrace{
\begin{pmatrix}
\vct{Z}_{n,K+1}\\
\vct{Z}_{n,K+2}\\
\vdots\\
\vct{Z}_{n,K+T}
\end{pmatrix}
}_{\vct{Z}_n}
\end{equation}
Here $a_{n',k} = \prod_{\ell\in[K+T]\backslash\{k\}}\frac{\alpha_{n'}-\beta_{\ell}}{\beta_{k}-\beta_{\ell}}$, for all $k \in [K+T]$.

As $\vct{Z}_n$ is uniformly random in $\mathbb{F}_p^{\frac{TM}{K}\times d_n D_n}$ and the matrix $\vct{A}_2$ comprised of Lagrange coefficients has full rank, $\vct{A}_2 \vct{Z}_n$ is also uniformly random in $\mathbb{F}_p^{\frac{TM}{K}\times d_n D_n}$. Now, for any $\vct{M} \in \mathbb{F}_p^{\frac{TM}{K} \times d_n D_n}$ and $\vct{N} \in \mathbb{F}_p^{M \times d_n D_n}$, we have
\begin{align}
    \textup{Pr}[\widetilde{\vct{X}}_{n,T}=\vct{M}|\overline{\vct{X}}_n=\vct{N}] &= \textup{Pr}[\vct{A}_2 \vct{Z}_n=\vct{M}-\vct{A}_1 \vct{N}|\overline{\vct{X}}_n=\vct{N}]\\
    & \overset{(a)}{=}\textup{Pr}[\vct{A}_2 \vct{Z}_n=\vct{M}-\vct{A}_1 \vct{N}]\\
    & \overset{(b)}{=}\frac{1}{p^{\frac{TMd_n D_n}{K}}}.
\end{align}
Here $(a)$ is because that $\overline{\vct{X}}_n$ and $\vct{Z}_n$ are independent, and $(b)$ is because that $\vct{A}_2 \vct{Z}_n$ is uniformly random in $\mathbb{F}_p^{\frac{TM}{K}\times d_n D_n}$. Next, we have 
\begin{align}
    \textup{Pr}[\widetilde{\vct{X}}_{n,T}=\vct{M}] = \sum_{\vct{N}} \textup{Pr}[\widetilde{\vct{X}}_{n,T}=\vct{M}|\overline{\vct{X}}_n=\vct{N}]\textup{Pr}[\overline{\vct{X}}_n=\vct{N}]\\
    =\sum_{\vct{N}} \frac{1}{p^{\frac{TMd_n D_n}{K}}} \textup{Pr}[\overline{\vct{X}}_n=\vct{N}] = \frac{1}{p^{\frac{TMd_n D_n}{K}}}.
\end{align}

We have from the above that for any $\vct{M}$ and $\vct{N}$, $\textup{Pr}[\widetilde{\vct{X}}_{n,T}=\vct{M}|\overline{\vct{X}}_{n}=\vct{N}] = \textup{Pr}[\widetilde{\vct{X}}_{n,T}=\vct{M}] = \frac{1}{p^{\frac{TMd_n D_n}{K}}}$, and hence $\widetilde{\vct{X}}_{n,T}$ and $\overline{\vct{X}}_{n}$ are statistically independent. That is, the mutual information $I\left(\widetilde{\vct{X}}_{n,T};\overline{\vct{X}}_{n}\right)=0$. As this holds for all $n > T$, we have $I(\widetilde{\vct{X}}_{T+1,T},\widetilde{\vct{X}}_{T+2,T},\ldots, \widetilde{\vct{X}}_{N,T}; \overline{\vct{X}}_{T+1}, \overline{\vct{X}}_{T+2}, \ldots, \overline{\vct{X}}_{N}) = 0$.

\section{Proof of Theorem \ref{thm4} }\label{sec:convergence_proof}
From Lemma 10 in \cite{NEURIPS2020_d714d2c5}, we state the following lemma that bounds the difference between the gradients of the losses computed from a sampled batch and all training data.
\begin{lemma}[]\label{lm2}
Consider mini-batch function $\nabla_n F_{\cal B}(\vct{W})\in\mathbb{R}^{p_n},n\in \{0,1,\ldots,N\}$, which satisfies $\mathbb{E}[\nabla_nF_{\cal B}(\vct{W})] = \nabla_n F(\vct{W})$. For $\epsilon <2L$, we have with probability at least $1-\delta$ that:
\begin{equation}
        \|\nabla_n F_{\cal B}(\vct{W}) - \nabla_n F(\vct{W})\|^2 \leq \frac{32 L^2(\log(2p_n/\delta)+\frac{1}{4})}{|\cal B|}.
\end{equation}
\end{lemma}
\begin{proof}
The proof refers to \cite{NEURIPS2020_d714d2c5}.
\end{proof}

Then we provide lemma \ref{lm3} to bound the loss function in each round $r$ as follows:
\begin{lemma}\label{lm3}
Under Assumption 2, for each round $r$, it follows that
\begin{equation}
     F(\vct{W}^{r+1}) \leq  F(\vct{W}^{r})+ \sum_{n=0}^N(L \eta_n^2-\frac{3}{2}\eta_n)\|\nabla_n F(\vct{W}^r)\|^2 
    + \sum_{n=0}^N L \eta_n^2\|\nabla_n F_{\cal B}(\widehat{\vct{W}}^r) -  \nabla_n F(\vct{W}^r)\|^2,
\end{equation}
\end{lemma}
\begin{proof}
From Assumption 1,
we can derive that:
\begin{equation}{\label{func1}}
    \begin{split}
    F(\vct{W}^{r+1}) =& F\left(\vct{W}_0^{r} - \eta_0\nabla_0 F_{\cal B}(\widehat{\vct{W}}^{r}) ,\ldots,\vct{W}_N^{r} - \eta_N\nabla_N F_{\cal B}(\widehat{\vct{W}}^{r}) \right)\\
    \leq& F(\vct{W}^r) - \sum_{n=0}^N\langle \nabla_n F(\vct{W}^r), \eta_n(\nabla_n F_{\cal B}(\widehat{\vct{W}}^{r})- \nabla_n F(\vct{W}^r) + \nabla_n F(\vct{W}^r)) \rangle 
+ \sum_{n=0}^N\frac{L}{2} \eta_n^2 \|\nabla_n F_{\cal B}(\widehat{\vct{W}}^r) \|^2 \\
   =& F(\vct{W}^r) - \sum_{n=0}^N\eta_n\|\nabla_n F(\vct{W}^r)\|^2 -  \sum_{n=0}^N\eta_n \langle \nabla_n F(\vct{W}^r), (\nabla_n F_{\cal B}(\widehat{\vct{W}}^{r})-\nabla_n F(\vct{W}^r))\rangle\\
   & + \sum_{n=0}^N\frac{L}{2} \eta_n^2 \|\nabla_n F_{\cal B}(\widehat{\vct{W}}^r)\|^2.
    \end{split}
\end{equation}

Note that we have:
\begin{equation}\label{func17}
\begin{split}
\|\nabla_n F_{\cal B}(\widehat{\vct{W}}^r)\|^2 =& \|\nabla _nF_{\cal B}(\widehat{\vct{W}}^r) - \nabla_n F(\vct{W}^r)+\nabla_n F(\vct{W}^r)\|^2 \\
 =& \|\nabla_n F_{\cal B}(\widehat{\vct{W}}^r) -  \nabla_n F(\vct{W}^r)\|^2 + \| \nabla_n F(\vct{W}^r)\|^2  \\
 & +2\langle \nabla_n F (\vct{W}^r),  \nabla_n F_{\cal B} (\widehat{\vct{W}}^r) - \nabla_n F(\vct{W}^r) \rangle,\forall n\in[N] .
\end{split}
\end{equation}

Combining (\ref{func1}) and (\ref{func17}), we have: 
\begin{equation}\label{ineq1}
\begin{split}
    F(\vct{W}^{r+1}) &\leq F(\vct{W}^{r})+ \sum_{n=0}^N(L \eta_n^2-\frac{3}{2}\eta_n)\|\nabla _nF(\vct{W}^r)\|^2 +\sum_{n=0}^N(L \eta_n^2-\frac{1}{2}\eta_n)\|\nabla_n F_{\cal B}(\widehat{\vct{W}}^r) -  \nabla_n F(\vct{W}^r)\|^2\\
    &\leq F(\vct{W}^{r})+ \sum_{n=0}^N(L \eta_n^2-\frac{3}{2}\eta_n)\|\nabla_n F(\vct{W}^r)\|^2 + \sum_{n=0}^NL \eta_n^2\|\nabla_n F_{\cal B}(\widehat{\vct{W}}^r) -  \nabla_n F(\vct{W}^r)\|^2,
\end{split}
\end{equation}
which completes the proof of lemma \ref{lm3}.
\end{proof}

\textbf{Proof of Theorem \ref{thm4}:}

Considering the stochastic rounding on clients' model parameters, from Assumption 1, 2, 4 and Lemma \ref{lm2}, we can derive the following inequality with probability at least $1-\delta$,  $\forall n\in \{0,1,\ldots,N\}$:
\begin{equation}\label{func20}
    \begin{split}
        \|\nabla_n F_{\cal B}(\widehat{\vct{W}}^r) - \nabla_n F(\vct{W}^k)\|^2 &\leq 2\|\nabla_n F_{\cal B}(\widehat{\vct{W}}^r)- \nabla_n F_{\cal B}(\vct{W}^r)\|^2 
        +2\|\nabla_n F_{\cal B}(\vct{W}^r) - \nabla_n F(\vct{W}^r)\|^2\\
        &\leq 2L^2\|(\vct{W}_0^r, Q_{stoc}(\vct{W}_1^r,\ldots, \vct{W}_N^r)) - \vct{W}^r\|^2 + 2\|\nabla_n F_{\cal B}(\vct{W}^r)- \nabla_n F(\vct{W}^r)\|^2 \\
        &\leq 2L^2\|Q_{stoc}(\vct{W}_0^r, \vct{W}_1^r,\ldots, \vct{W}_N^r) - \vct{W}^r\|^2 + 2\|\nabla_n F_{\cal B}(\vct{W}^r)- \nabla_n F(\vct{W}^r)\|^2\\
        & = 2L^2\gamma^2\sigma^2 + \frac{64 L^2(\log(2p_n/\delta)+\frac{1}{4})}{|\cal B|}\\
        & = 2L^2\gamma^2\sigma^2 + 2V_n,
    \end{split}
\end{equation}
where $V_n = \frac{32 L^2(\log(2p_n/\delta)+\frac{1}{4})}{|\cal B|}$.

When $\eta_n\leq \frac{3}{4L}, \forall n\in \{0,1,\ldots,N\}$ in Lemma \ref{lm3}, the following inequality holds:
\begin{equation}
\label{func7}
   \sum_{n=0}^N\frac{3}{4}\eta_n \|\nabla_n F(\vct{W}^r)\|^2 \leq  F(\vct{W}^{r}) - F(\vct{W}^{r+1}) 
    + \sum_{n=0}^N L \eta_n^2\|\nabla_n F_{\cal B}(\widehat{\vct{W}}^r) -  \nabla_n F(\vct{W}^r)\|^2,
\end{equation}

Under Assumption 3, taking expectation on both sides of (\ref{func7}) and adopting $\eta_n = \frac{3}{4L}\sqrt{\frac{1}{R}} \leq \frac{3}{4L},\forall n\in \{0,1,\ldots,N\}$, the following holds with probability at least $1-\delta$:
\begin{equation}
    \begin{split}
        \frac{1}{R}\sum_{r=0}^{R-1}\mathbb{E}\left(\sum_{n=0}^N\|\nabla_n F(\vct{W}^r)\|^2\right) &\leq \frac{ F(\vct{W}^{0}) - F(\vct{W}^{*})}{\frac{3}{4}\eta_n R} +\frac{\sum_{r=0}^{R-1} \mathbb{E}(\sum_{n=0}^NL\eta_n^2\|\nabla_n F_{\cal B}(\widehat{\vct{W}}^r) -  \nabla_n F(\vct{W}^r)\|^2)]}{\frac{3}{4}\eta_n R}\\
        &\leq \frac{16L}{9\sqrt{R}} (F(\vct{W}^{0}) - F(\vct{W}^{*}))+ \frac{\sum_{n=0}^N(2L^2\gamma^2\sigma^2 + 2V_n )}{\sqrt{R}}\\
        &= \mathcal{O}\left(\frac{1}{\sqrt{R}}\right).\\
    \end{split}
\end{equation}
This completes the proof of Theorem \ref{thm4}.

\section{Six datasets' descriptions, models and training details} \label{sec:dataset}

\begin{table*}[ht]
\caption{Dataset Descriptions.}
\label{sample-table}
\vskip 0.15in
\scriptsize
   \centering
   \resizebox{1.0\columnwidth}{!}{
    \begin{tabular}{c c c c c c c }
    \hline
     & \multicolumn{2}{c}{Tabular} &\multicolumn{2}{c}{Multi-view} &\multicolumn{2}{c}{CV} \\
      \cmidrule(lr){2-3} \cmidrule(lr){4-5} \cmidrule(lr){6-7}
          & Parkinson & Credit card & Handwritten  & Caltech-7 & EMNIST &FashionMNIST  \\
       \hline 
         Number of samples & 756 & 30,000 & 2000 & 1474 &131,600 & 70,000  \\
         \hline
         Feature size  & 754 & 24 &  649&3766 & 784 &784  \\
         \hline
         Number of classes & 2& 2&10 & 7 & 47 & 10  \\
         \hline
    \end{tabular}}
    \label{tab:my_label}
\end{table*}

\textbf{Parkinson:} The dataset's features are biomedical voice measurements from 31 people, 23 with Parkinson's disease (PD). Each feature is a particular voice measurement. The label is divided to 0 and 1, which represents PD and healthy people. There are 10 clients with vertically partitioned data. 70\% of the data is regarded as training data and the remaining part is test data. Each client holds a 2-layer PN, and the server holds a network with 2 Linear-ReLU layers and 1 Linear-Sigmoid layer. The learning rate is set to 0.005. The batch size is 16.

\textbf{Credit Card:} The dataset is composed of information on default payments, demographic factors, credit data, history of payment, and bill statements of credit card clients in Taiwan from April 2005 to September 2005, which contains 24 attributes. The labels of the samples are biased since there are 78\% of samples labeled as 0 and 22\% of samples labeled as 1 to denote default payment. 11 clients equally hold vertically partitioned data. Each client holds a 2-layer PN, and the server holds a network with a Linear-BatchNorm-Linear-ReLU-BatchNorm-WeightNorm-Linear-Sigmoid structure. The learning rate is set to 0.01 and the batch size is set to 32.

\textbf{FashionMNIST:} 
It is an image dataset related to household goods. Each image sample is evenly partitioned across 28 clients. 
Each client holds a 1-layer PN, and the server holds a network with 2 Linear-ReLU layers and one Linear-Logsoftmax layer. The learning rate is set to 0.05 and batch size 256 is selected.

\textbf{EMNIST:} The EMNIST dataset is a set of handwritten character digits converted to a 28x28 pixel image format. There are 28 clients and the data is partitioned the same as FashionMNIST. The clients' model is a 2-layer PN, and the server model is the same as the above FashionMNIST server's model. The learning rate is 0.05 and the batch size is 256.

\textbf{HandWritten:} The dataset consists of features of handwritten numerals 0-9 extracted from a collection of Dutch utility maps. 200 patterns per class (for a total of 2,000 patterns) have been digitized in binary images. It consists of 6 views, pixel (PIX) of dimension 240, Fourier coefficients of dimension 76, profile correlations (FAC) of dimension 216, Zernike moments (ZER) of dimension 47, Karhunen-Loeve coefficients (KAR) of dimension 64 and morphological features (MOR) of dimension 6. Each client holds one view. The dataset is split to 60\% as the train set and 40\% as the test set. Each client holds a 2-layer PN, and the server holds a model with 2 Linear-ReLU layers and 1 Linear-Logsoftmax layer. The learning rate is 0.02 and the batch size is 32.

\textbf{Caltech-7:} Caltech-101 is an object recognition dataset containing 8677 images of 101 categories. 7 classes of Caltech 101 are selected, i.e., Face, Motorbikes, Dolla-Bill, Garfield, Snoopy, Stop-Signand Windsor-Chair. The dataset is composed of 6 views, each of which is held by a client. 80\% of the dataset is used for training and 20\% for testing. Each client holds a 2-layer PN. The server holds the same model structure as HandWritten. The learning rate is 0.01 and the batch size is 8.